\documentclass[review]{elsarticle}
\usepackage{lineno,hyperref}
\usepackage{algorithmic}
\usepackage{algorithm}
\usepackage[table,xcdraw]{xcolor}
\usepackage{graphicx, caption}
\usepackage{booktabs}
\usepackage{longtable}
\graphicspath{ {fig/} }
\usepackage{tabularx}
\modulolinenumbers[5]
\journal{Journal of Pattern Recognition}

\begin{document}

\begin{frontmatter}

\title{Similarity Based Clustering for Enhancing Image Classification Architectures}


\author[1]{Dishant Parikh\corref{cor1}}
\ead{dishant30899@gmail.com}
\cortext[cor1]{Corresponding author}
\address[1]{Independent researcher}

\begin{abstract}
Convolutional networks are at the center of best-in-class computer vision applications for a wide assortment of undertakings. Since 2014, a profound amount of work began to make better convolutional architectures, yielding generous additions in different benchmarks. Albeit expanded model size and computational cost will, in general, mean prompt quality increases for most undertakings but, the architectures now need to have some additional information to increase the performance. I show evidence that with the amalgamation of content-based image similarity and deep learning models, we can provide the flow of information which can be used in making clustered learning possible. The paper shows how training of sub-dataset clusters not only reduces the cost of computation but also increases the speed of evaluating and tuning a model on the given dataset. 

\end{abstract}


\end{frontmatter}

\section{Impact statement}
With SBC-ICA, researchers can work with the following advantages:
\begin{enumerate}
    \item Evaluate models quickly. 
    \item Faster convergence. 
    \item Train with lower computational resources.
    \item Have an ability to build human-induced conditional model.
    \item Faster extensions. 
\end{enumerate}

\section{Introduction}
Understanding the world in a solitary look is one of the most practiced accomplishments of the human mind. It takes only milliseconds to perceive the classification of an object or an action, underlining a significant job of feedforward processing in visual analysis. With the introduction of deep convolutional neural networks \cite{huang2017densely}, which achieved breakthrough accuracies in the domain of image classifications, there has been a profuse amount of work in developing better architectures and training methodologies \cite{szegedy2017inception}, \cite{he2016deep}. The current work, in image classification, is more focused on the networks and lesser when it comes to the understanding of the dataset and its structure. 

For instance, we saw the work on depthwise separable network architectures \cite{howard2017mobilenets}, \cite{sandler2018mobilenetv2} and dynamic model scaling \cite{tan2019efficientnet}, with the methodologies like the Noisy student’s semi-supervised learning \cite{xie2020self}, which brought a higher accuracy on ImageNet while providing an intuition into self-learning and distillation. Although structures like these have produced astonishing results on many of the large databases like ImageNet \cite{krizhevsky2012imagenet}, they still need to be tweaked and trained in a particular way to get the best of object detection \cite{bianco2018benchmark}, while dealing with fine-grained datasets \cite{khosla2011novel}. We have seen a significant improvement in accuracy by tuning the models based on knowledge acquired from studying the structure of the dataset. So what if we append that type of information to the model itself? 

The answer came from a field that has yet not been amalgamated, with modern deep learning architectures: Image similarity \cite{wang2014learning}. Since the foundation of the field, we have seen a ton of work in understanding how machines see an image. For example, when image fusion evolved with the idea to make the output image show the understanding of the scene more elaborately than all the input images before, it was a sincere thought into the development of image retrieval systems \cite{ciocca2001content}, \cite{liu2007survey}, \cite{long2003fundamentals}, \cite{wan2014deep}. And by combining the use of deep learning to understand the content-based image similarities, it was now easier to develop accurate analysis models. We show how by amalgamating these two fields, we achieve even better architectures. 

I show the need for an additional meta-data field, which I call: Similarity Matrix. It could help save hours of computational time and also increase the accuracy of the deepest of the architectures on class-similar datasets. While working with the datasets which contain a hodgepodge of similar classes and distinctive ones, for example, Stanford’s dog breeds dataset \cite{khosla2011novel}, it is difficult to tune the model efficiently for better performance. The problem is not only the drop in accuracy but also the overfitting of the model. The standard techniques could only provide a way out of overfitting up to a certain limit \cite{srivastava2014dropout}. 

\subsection{Clustered training}
What I propose is a clustering of the dataset into two or more sub-datasets, with sequential training to improve existing architectures. The clusters are formed based on the similarity matrix of all the classes inside the dataset, and hence we find the most similar classes together in one cluster. The main reason for grouping similar classes together was to get a little lower number of acceptable clusters which have enough intercluster distance for the master classifier to accurately select an appropriate cluster for the new image inference. 

Many would argue that while clustering, it is better to group dissimilar classes to ensure even better accuracy. This idea does make sense for a lower number of classes. But when we go even beyond 20 classes, the scalability issues are strongly reflected. Let us assume that there are 50 classes in a dataset. Realistically speaking, how many clusters would need to be formed so that clustering dissimilar classes help the classification accuracy? Also, how far would these clusters be from each other, i.e., the intercluster distance, so that the master classifier (the one that chooses the appropriate cluster for the inference image to go to) will be accurate? Here are the two possibilities:

\begin{enumerate}
    \item If the number of clusters is low, the classification accuracy inside a particular cluster will be as good as an accuracy on the entire dataset. Additionally, the master classifier will not be able to classify it too well as there is no real set of distinctiveness between clusters. The cluster choosing accuracy would be too low. 
    \item If the number of clusters is high, the same problem of choosing the correct cluster appears, with the additional cost in training (as there might not be that much parallelism available) 
\end{enumerate}

Both of these options are in no way better than training on the master dataset directly.
When we group similar classes, the intercluster distance is quite high as the clusters will have those distinctive features we talked about earlier. And when the number of classes to classify is lower, the classification accuracy will anyway tend to go higher as well, as better architectures specific to that cluster can be used to improve it. Due to these reasons, the grouping of similar classes together is a better choice.

Once we generate the sub-dataset clusters, we train independent models on each sub-dataset. At inference, we use the feature vectors to choose the best prediction model. This way, we reduce the computational needs as well as retain (and in some cases increase) the model accuracies, even with a limited number of epochs.

\section{Related works}
Until recently, most of the work in Image classification has been done, by making the models deeper and the labeled datasets larger. The recent work, in dynamic model scaling, did show the intent of tweaking the model by resolution, depth, and width, which can make it efficient without making it unnecessarily complicated \cite{lu2007survey}. Researchers have also shown much work in the model architecture’s efficiency, like introducing inverted residuals and linear bottlenecks for lighter models \cite{sandler2018mobilenetv2} and descriptor pyramids \cite{iandola2014densenet}. Some work in the latest architecture revolved around the kernel functions \cite{liu2020image}. We also saw the work in Deep sparse rectifier networks \cite{glorot2011deep} and even in understanding the difficulties of training deeper networks \cite{glorot2010understanding}. We even saw some changes in base architectures to further improve performance \cite{szegedy2016rethinking}, but mostly we thought that the deeper architectures and faster computational machinery would provide the accuracies that we want. 

Even though possible, sometimes, the problem of overfitting and the cost of computing becomes a huge hurdle. Even with state-of-the-art models, we are not able to achieve impressive accuracy when it comes to class-similar datasets. What we tried is to check the possibility of improving the methodology by amalgamating the knowledge, driven by the image similarity of classes in the datasets. Much work has been done, in understanding the image, by using content-based analysis \cite{smeulders2000content} or even complex wavelet structural analysis \cite{sampat2009complex}. But the work was mostly transferred to image retrieval systems. We saw work on using this analysis for making the application better \cite{agarwal2015detection}, which was used not only on content-based image analysis but also object-based image analysis \cite{hoiem2004object}. Further, we saw the development of similarity engines \cite{sasaki2005formulation} and the best-in-class content-based image retrievals.
 
With the use of fundamentals of content-based image analysis \cite{long2003fundamentals}, we produce a similarity matrix for the entire dataset. The similarity between the classes does provide critical insight into training better models. Not improving the architecture, but improving how these architectures are trained. We hence developed a generic methodology of clustering the data into sub-datasets and then train the models independently on them. Once we get the set of models, the new query image needs to make a correct choice of model to send to for prediction. We provide the clustered-model training method for faster training and more accurate models for class-similar datasets.

\section{Methodology}

\subsection{Dataset information}

\begin{enumerate}
    \item Stanford dogs dataset. 
    
    Stanford dog breeds dataset \cite{khosla2011novel} contains 20,580 images of 120 dog breeds from across the world. Each class contains roughly 150 images. I have used in total 6 versions of this dataset. The original data containing 20,580 images was split in two and three-cluster dataset. In the two cluster split, the first cluster contained 76 classes and the second had 44 classes. In the three cluster split, classes were distributed in the ratio of 33:22:5 across the three splits. The entire list of splits has been provided in Appendix B.
    
    \item Oxford flowers dataset.
    
    Oxford Flowers dataset \cite{nilsback2008automated} contains 8189 images of 102 flower species. Each class consists of between 40 and 258 images. I have used six versions of this dataset. Like in dogs dataset, the original dataset was split into two-cluster and three-cluster dataset. In the two cluster split, the first cluster contained 86 classes and the second had 16 classes. In the three cluster split, classes were distributed in the ratio of 16:69:17 across the three splits.

\end{enumerate}

\subsection{Feature Extraction}

The first step into the generation of the DSI is in understanding the machine’s perspective, .i.e., features that the machine extracts. For the same, I use a pretrained ResNet152 and make a forward pass on all the images. The feature vector is extracted by taking the vector before it is passed to the average pooling layer. Hence, the length of the feature vector, in the case of ResNet152, will be 2048. The reason for extracting the features from the last layer is because the latter layers have more specific features while the previous ones have more general features \cite{yosinski2014transferable}. For feature extraction there are two other layers available, one is passing the vector to max-pooling and then extracting it. The second is to take the vector at the fully connected layer, which in our case will be 1000 units long. But according to the work by \cite{Pohchih2020}, the features work well in the case of Avg. pooling especially while working with content-based image similarity. 

Another case to note is the model selection. Although any model can be used for feature extraction, the choice will depend a lot on the depth of feature extraction required. I have noticed that in most cases, the specific features are better extracted with deeper extraction. The argument can also be supported with the results on classification algorithms comparison \cite{bressem2020comparing} as well as the comparative results in the model choice in CBIR systems \cite{Pohchih2020}.

\subsection{Similarity matrix generation}

There are three steps to similarity matrix generation. 
\begin{enumerate}
    \item Feature centroid generation.
    \item Calculating similarity by comparing the distances of feature centroids.
    \item Generating similarity matrix by placing the pairwise distance evaluations of the similarity for each class in the dataset.
\end{enumerate}

Feature centroids are generated by fitting an unsupervised clustering algorithm with one cluster and extracting the cluster centers as the feature centroids. The use of clustering algorithms also helps in checking the inertia, that is, the overall spread of the features in the class, which in turn can be used for checking the overall variety of the class. But this way of checking variety works only if there is no other factor to be taken into account. There are two reasons for computing feature centroids: One, reducing the number of comparisons required. Second, the class should not be checked based on either outliers or extremes, that will only hurt the distance computations for similarity.

Once the feature centroids are extracted, the similarity between the classes can be computed by taking the cosine distance (as per equation 1) between the feature centroids of each class. The reason for using cosine distance is that it achieves the highest mMAP score in CBIR systems when used with the feature vector of the average pooling layer. Additionally, according to the comparative study as well as a poll of 100 human volunteers, distances like Cosine, Euclidean, Manhattan, Vector Cosine Angle Distance (VCAD) achieved the same amount of contradiction between similarities calculated by machine and humans \cite{hasnat2013comparative}. Finally, the matrix is generated by computing the pairwise distances of all classes with all other classes in the dataset. The synopsis of the methodology can be viewed in Figure \ref{fig:figure1}.

\renewcommand{\thefigure}{1}
\begin{figure}[!t]
\centering
\includegraphics[width = 0.9\columnwidth,scale=0.2]{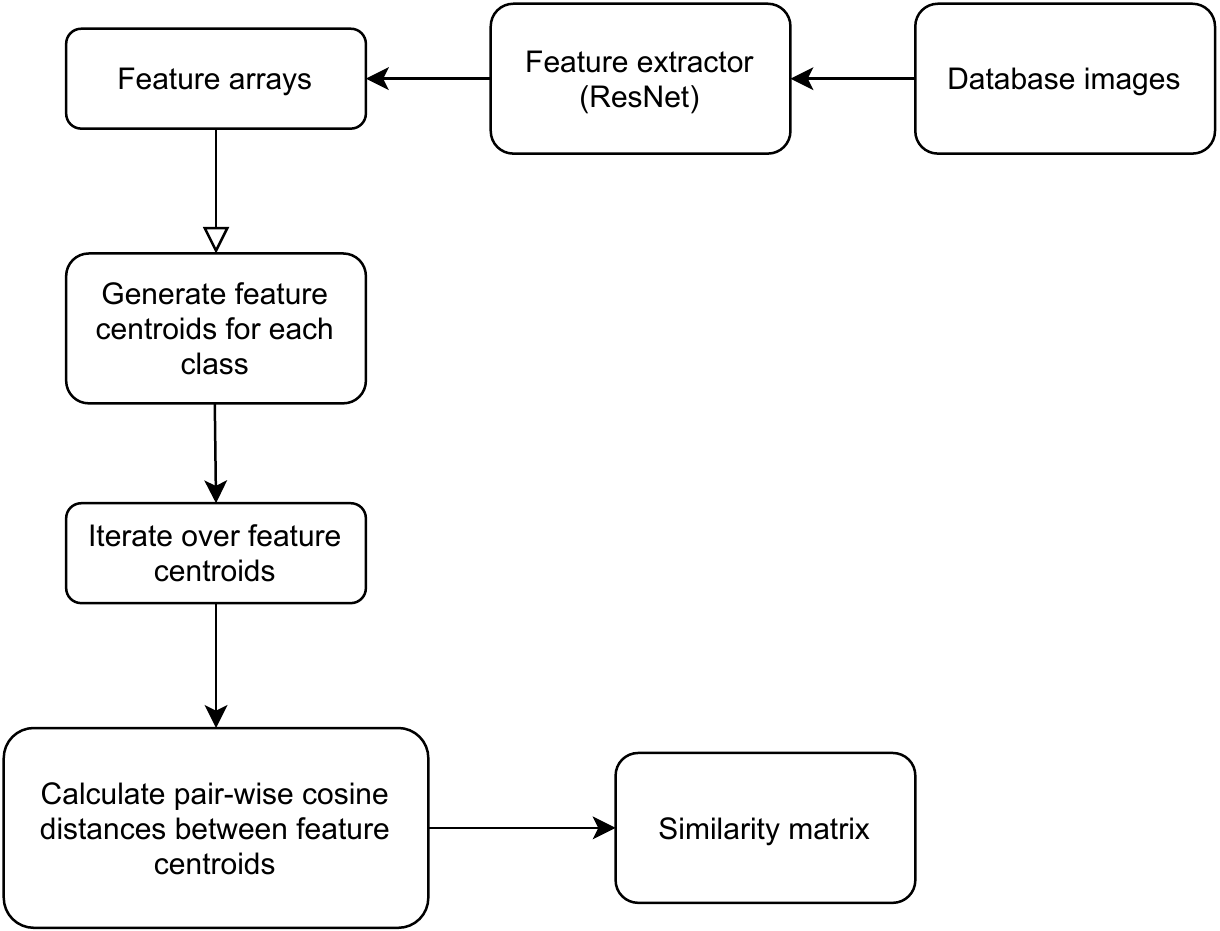}
\caption{Similarity matrix generation}
\label{fig:figure1}
\end{figure}

\renewcommand{\thefigure}{2}
\begin{figure}[!t]
\centering
\includegraphics[width = 0.9\columnwidth,scale=0.2]{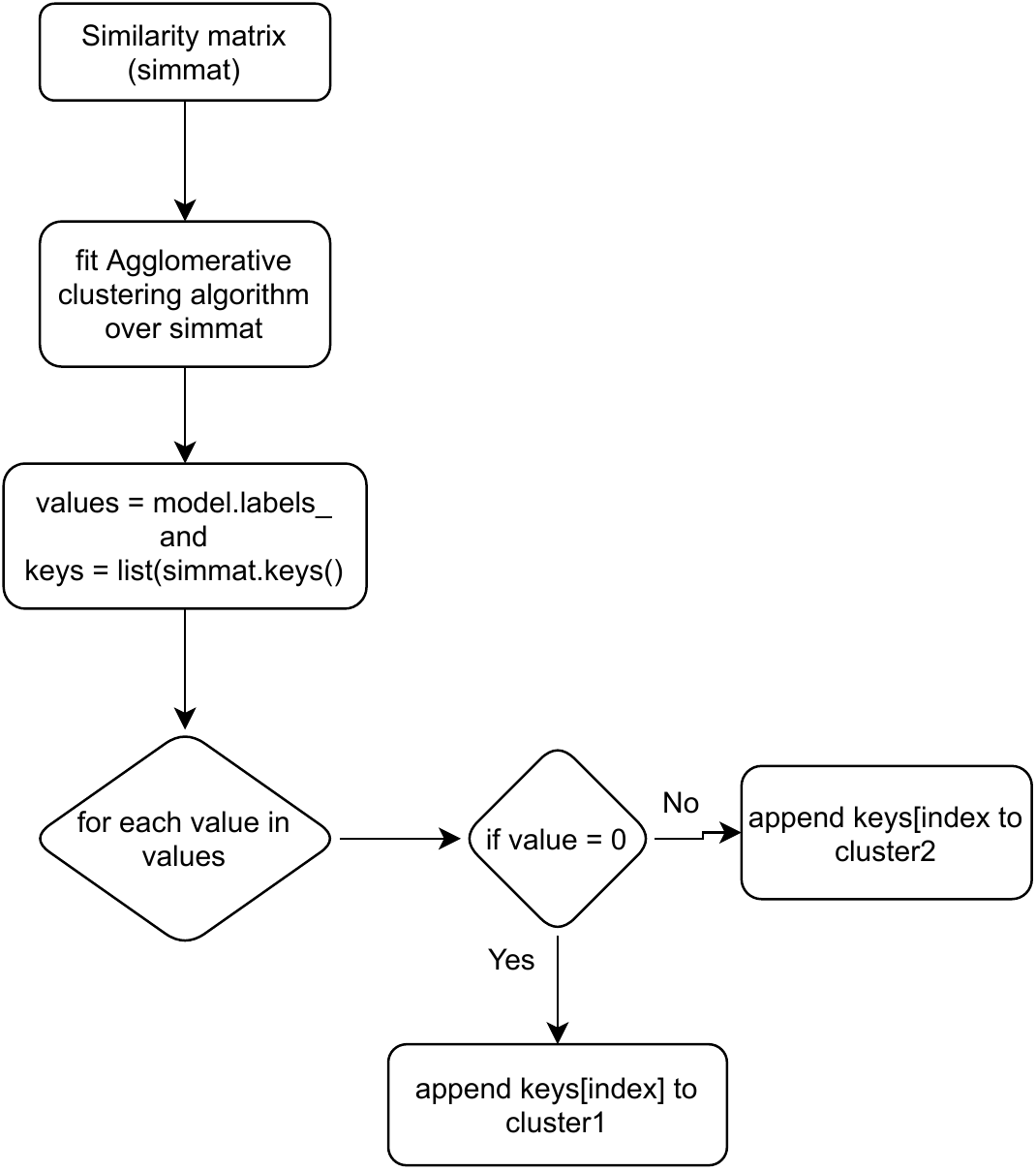}
\caption{Cluster formation technique}
\label{figure2}
\end{figure}

\subsection{Cluster formation technique}

The cluster formation technique uses the similarity matrix and the number of clusters as input. The algorithm then fits a hierarchical clustering algorithm over the similarity matrix with the number of clusters defined. The labels of the fit are used to form the two clusters. The use of hierarchical clustering was due to the ability to govern the decisions based on linkages rather than distance-based values, as the similarity matrix was already defined. I used all the default values of hierarchical clustering with Ward’s linkage. I used Ward’s method linkage because it outperforms the complete linkage when the clusters overlap, according to the comparative study by Vijaya \cite{vijaya2017review}. 

\renewcommand{\thefigure}{3}
\begin{figure}[!t]
\centering
\includegraphics[width = 0.9\columnwidth,scale=0.2]{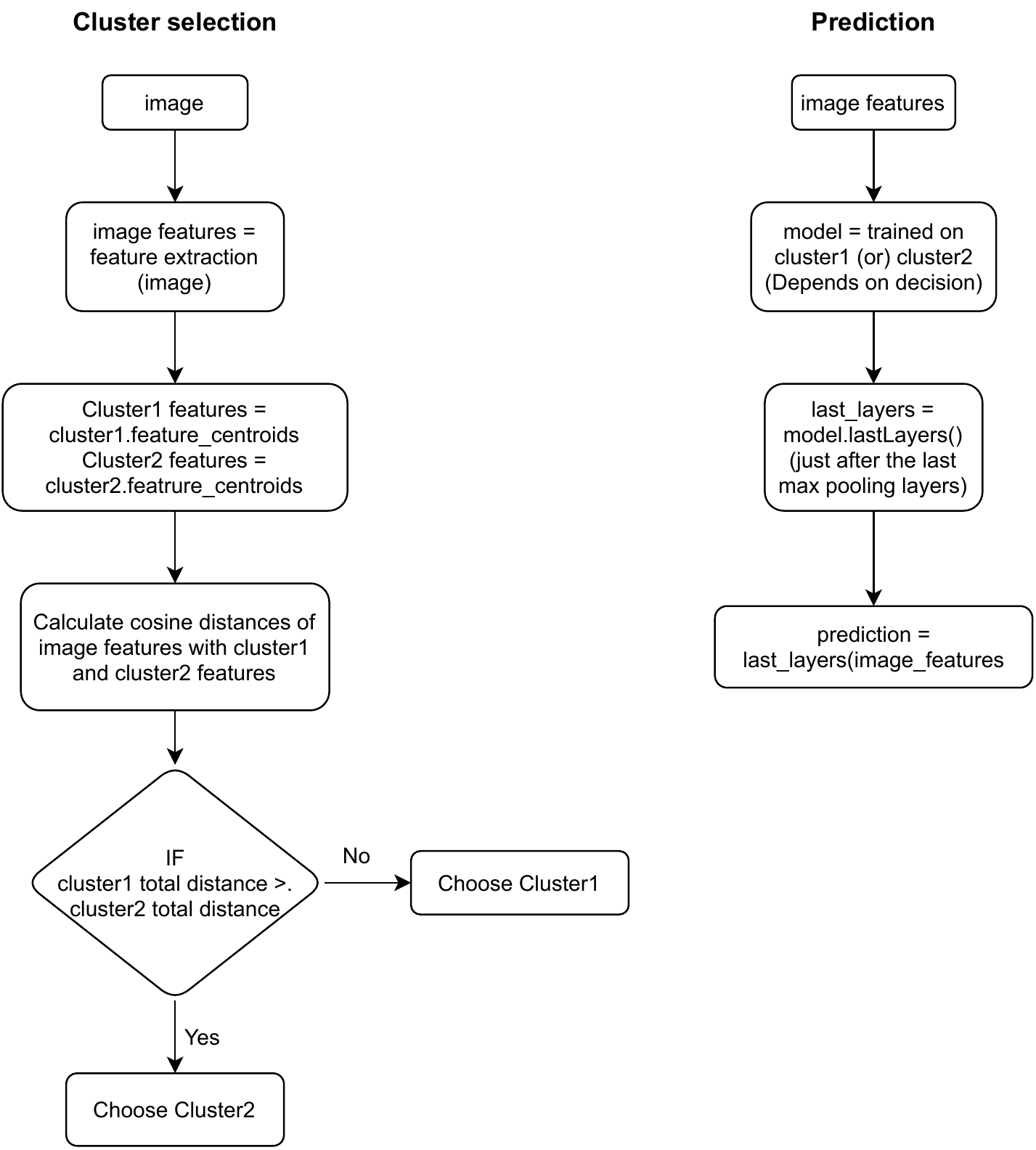}
\caption{Inference methodology}
\label{figure3}
\end{figure}

\subsection{Inference cluster selection technique}

At the time of inference, I run the feature extraction model to extract inference image features. The feature vector is compared with the feature centroids of both the clusters, using the same model which was used for clustering. It makes the cluster selection efficient. But, after the cluster selection step, there is no need to process the image again, as the features are already extracted. I directly used the feature vector as input to the fully connected layer and ran the forward pass. It makes the entire inference efficient, and the time taken remains the same. The inference methodology can be seen in Figure \ref{figure3}.

\subsection{Model architectures}

I have used 4 traditional vision model architectures to train upon all the datasets. The architectures are: VGG-19, ResNet50, AlexNet and ResNeXt50. All the models are pre-trained on ImageNet and use the standard configuration with Cross entropy loss and step learning rate scheduler.

\subsection{Training environment }
All the models were trained on Intel Xeon E5-1620 v3 (@ 3.50GHz × 8 CPU)  with Nvidia GeForce RTX2080Ti GPU and 16GB of RAM. I used the standard PyTorch library to code and train the model architectures. All models were trained with CUDA enabled. \footnote{All code can be accessed at-https://github.com/Dishant-P/Similarity-based-clustering-Official-research-module.git}


\begin{algorithm}
\textbf{INPUT:}\\$f{\_}list (n_{c} \times n_{i} \times 2048)$: List of feature arrays of every class image in the database\\
\textbf{OUTPUT:}\\$sim{\_}mat (n_{c}\times n_{c})$: Similarity matrix\\$n_{c}:$ number of classes\\$n_{i}:$ number of images per class\\

\begin{algorithmic}
\STATE \textbf{Function} Similarity{\_}matrix{\_}generator($f{\_}list$, $classes$):
\STATE \hspace{0.3175cm} $i \leftarrow 0$
\STATE \hspace{0.3175cm}\textbf{for} length of $f{\_}list$ do

\STATE \hspace{0.635cm}$class{\_}features \leftarrow features[classes[i]]$ 
\STATE \hspace{0.635cm}$model \leftarrow KMeans(n{\_}clusters = 1)$
\STATE \hspace{0.635cm}$model.fit(class{\_}features)$
\STATE \hspace{0.635cm}$feature{\_}centroids[classes[i]] \leftarrow model.cluster{\_}centers{\_}$
\STATE \hspace{0.635cm}$i \leftarrow i + 1$

\STATE \hspace{0.5cm}
\STATE \hspace{0.3175cm} $i \leftarrow 0$
\STATE \hspace{0.3175cm}\textbf{for} length of $feature{\_}centroids$ do
\STATE \hspace{0.635cm} $j \leftarrow 0$
\STATE \hspace{0.635cm}\textbf{for} length of $feature{\_}centroids$ do
\STATE \hspace{0.9525cm}$simmat[classes[i]] \leftarrow cosine(feature{\_}centroids[i],$
\STATE \hspace{5.27cm}$feature{\_}centroids[j])$
\STATE \hspace{0.9525cm}$j \leftarrow j + 1$
\STATE \hspace{0.635cm}$i \leftarrow i + 1$
\STATE \hspace{0.3175cm}\textbf{return} $simmat$

\end{algorithmic}
\caption{Algorithm to generate Similarity Matrix}
\label{algo:algo1p1}
\end{algorithm}


\begin{algorithm}
\textbf{INPUT:} $sim{\_}mat (n_{c}\times n_{c})$: Similarity matrix\\ $num{\_}clusters$: Number of clusters to be formed 
\\$n_{c}:$ number of classes\\
\textbf{OUTPUT:} Defining the sub-dataset cluster splits\\
\begin{algorithmic}
\STATE \textbf{Function} cluster{\_}formations($simmat, num{\_}clusters$):
\STATE \hspace{0.3175cm}$model \leftarrow HierarchicalClustering(n{\_}clusters = num{\_}clusters)$
\STATE \hspace{0.3175cm}$model.fit(simmat)$
\STATE \hspace{0.3175cm}$values \leftarrow model.labels{\_}$
\STATE \hspace{0.3175cm}$keys \leftarrow list(simmat.keys())$
\STATE \hspace{0.3175cm} $i \leftarrow 0$
\STATE \hspace{0.3175cm}\textbf{for} length of $values$ do
\STATE \hspace{0.635cm} {\#} In case of num{\_}clusters = 2
\STATE \hspace{0.635cm}\textbf{if} $values[i] = 0$
\STATE \hspace{0.9525cm}$cluster1.append(keys[i])$
\STATE \hspace{0.635cm}\textbf{else}
\STATE \hspace{0.9525cm}$cluster2.append(keys[index])$
\STATE \hspace{0.635cm} $i \leftarrow i + 1$
\end{algorithmic}
\caption{Algorithm for defining clusters}
\label{algo:algo2}
\end{algorithm}


\begin{algorithm}
\textbf{INPUT:} $image$: The inference image\\
\textbf{OUTPUT:} Prediction
\begin{algorithmic}
\STATE \textbf{Function} select{\_}cluster(image)
\STATE \hspace{0.3175cm} {\#} In case of two clusters
\STATE \hspace{0.3175cm}$cluster1{\_}features \leftarrow cluster1.feature{\_}centroids$
\STATE \hspace{0.3175cm}$cluster2{\_}features \leftarrow cluster2.feature{\_}centroids$
\STATE \hspace{0.3175cm}$image{\_}features \leftarrow feature{\_}extraction(image)$
\STATE \hspace{0.3175cm}$c1{\_}score \leftarrow 0$
\STATE \hspace{0.3175cm}$c2{\_}score \leftarrow 0$
\STATE \hspace{0.3175cm} $i \leftarrow 0$
\STATE \hspace{0.3175cm}\textbf{for} length of $cluster1{\_}features$ do
\STATE \hspace{0.635cm}$c1{\_}score \leftarrow c1{\_}score + cosine(image{\_}features, cluster1{\_}features[i])$ 
\STATE \hspace{0.635cm}$i \leftarrow i + 1$
\STATE \hspace{0.3175cm} $i \leftarrow 0$
\STATE \hspace{0.3175cm}\textbf{for} length of $cluster2{\_}features$ do
\STATE \hspace{0.635cm}$c2{\_}score \leftarrow c2{\_}score + cosine(image{\_}features, cluster2{\_}features[i])$
\STATE \hspace{0.635cm}$i \leftarrow i + 1$
\STATE \hspace{0.3175cm}\textbf{if} $c1{\_}score > c2{\_}score$
\STATE \hspace{0.635cm} \textbf{return} $1, image{\_}features$
\STATE \hspace{0.3175cm}\textbf{else}
\STATE \hspace{0.635cm} \textbf{return} $0, image{\_}features$
\end{algorithmic}


\begin{algorithmic}
\STATE \textbf{Function} predict{\_}class(image)
\STATE \hspace{0.3175cm}$cluster1{\_}decision, features \leftarrow select{\_}cluster(image)$
\STATE \hspace{0.3175cm}\textbf{if} $cluster{\_}decision = 0$
\STATE \hspace{0.635cm}$model \leftarrow trained{\_}on{\_}cluster1()$
\STATE \hspace{0.3175cm}\textbf{else}
\STATE \hspace{0.635cm}$model \leftarrow trained{\_}on{\_}cluster2()$
\STATE \hspace{0.3175cm}$last{\_}layers \leftarrow model.lastLayers()$
\STATE \hspace{0.3175cm}$prediction \leftarrow last{\_}layers(features)$
\STATE \hspace{0.3175cm}\textbf{return} $prediction$

\end{algorithmic}

\caption{Inference selection technique}
\label{algo:algo3}
\end{algorithm}

\section{Experiments and discussions}
As we break down the dataset into individual clusters that can be trained upon independently, it makes it easier for us to evaluate any change in model architecture. As similar classes would pose the most difficult task, the accuracy achieved in the cluster dataset, will in turn be reflected in the entire dataset trained model. Due to this, it is possible to experiment and evaluate different methodologies in ¼th or lower time. With this, it is also possible to use different models for different sub-dataset clusters. Due to this, we can make sure to use the best possible model for that particular cluster. It makes a significant difference not only in the speed of model training but also in the accuracy.

\begin{figure}[!t]
\centering
\begin{minipage}{.5\textwidth}
    \renewcommand{\thefigure}{4}
    \centering
    \captionsetup{width = 0.9\columnwidth}
    \includegraphics[width = 0.9\columnwidth,scale=0.2]{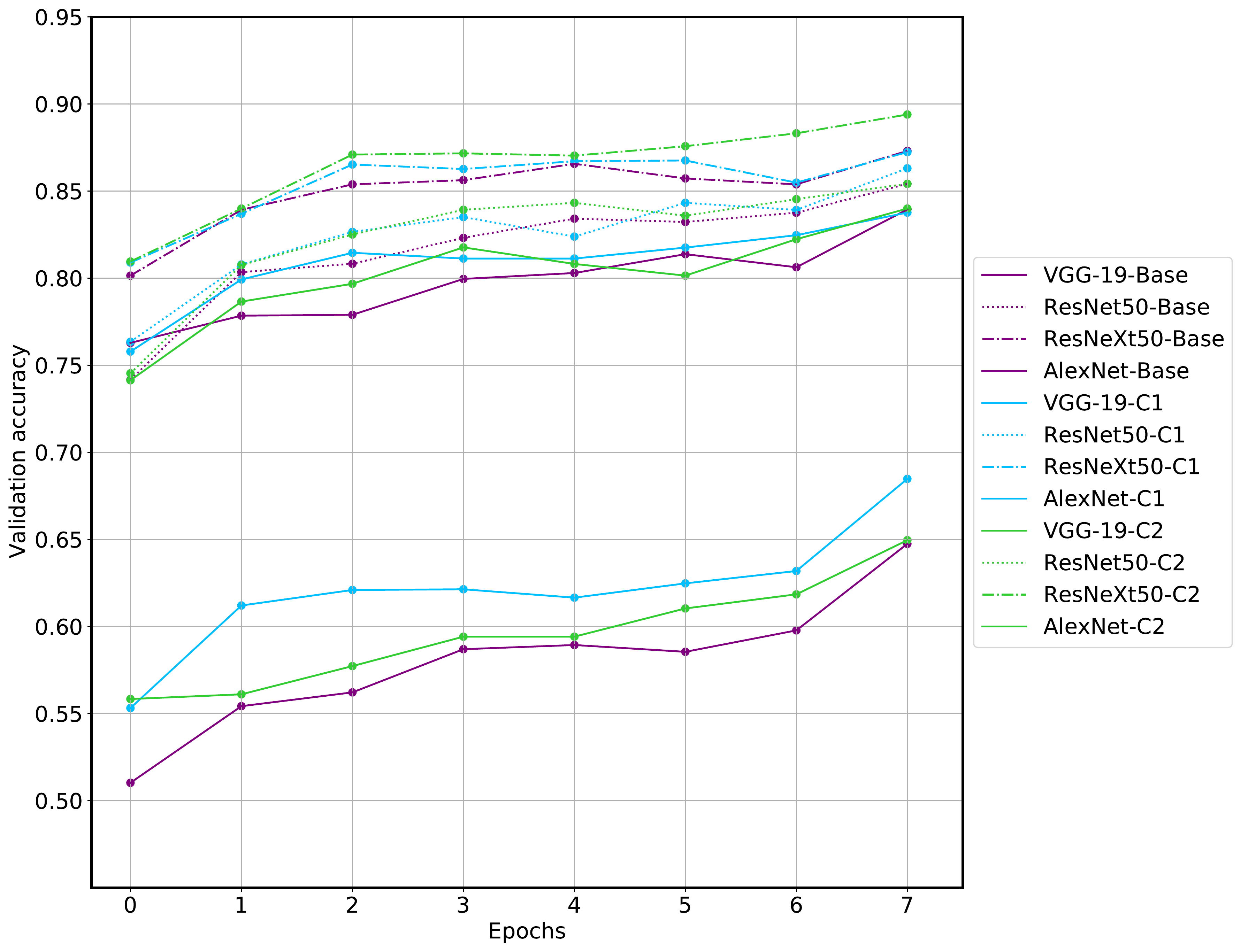}
    \caption{Validation accuracies for each epoch for model architectures trained on two cluster splits and original Stanford dogs dataset.}
    \label{figure4}
\end{minipage}%
\begin{minipage}{.5\textwidth}
    \renewcommand{\thefigure}{5}
    \centering
    \captionsetup{width = 0.8\columnwidth}
    \includegraphics[width = 0.9\columnwidth,scale=0.2]{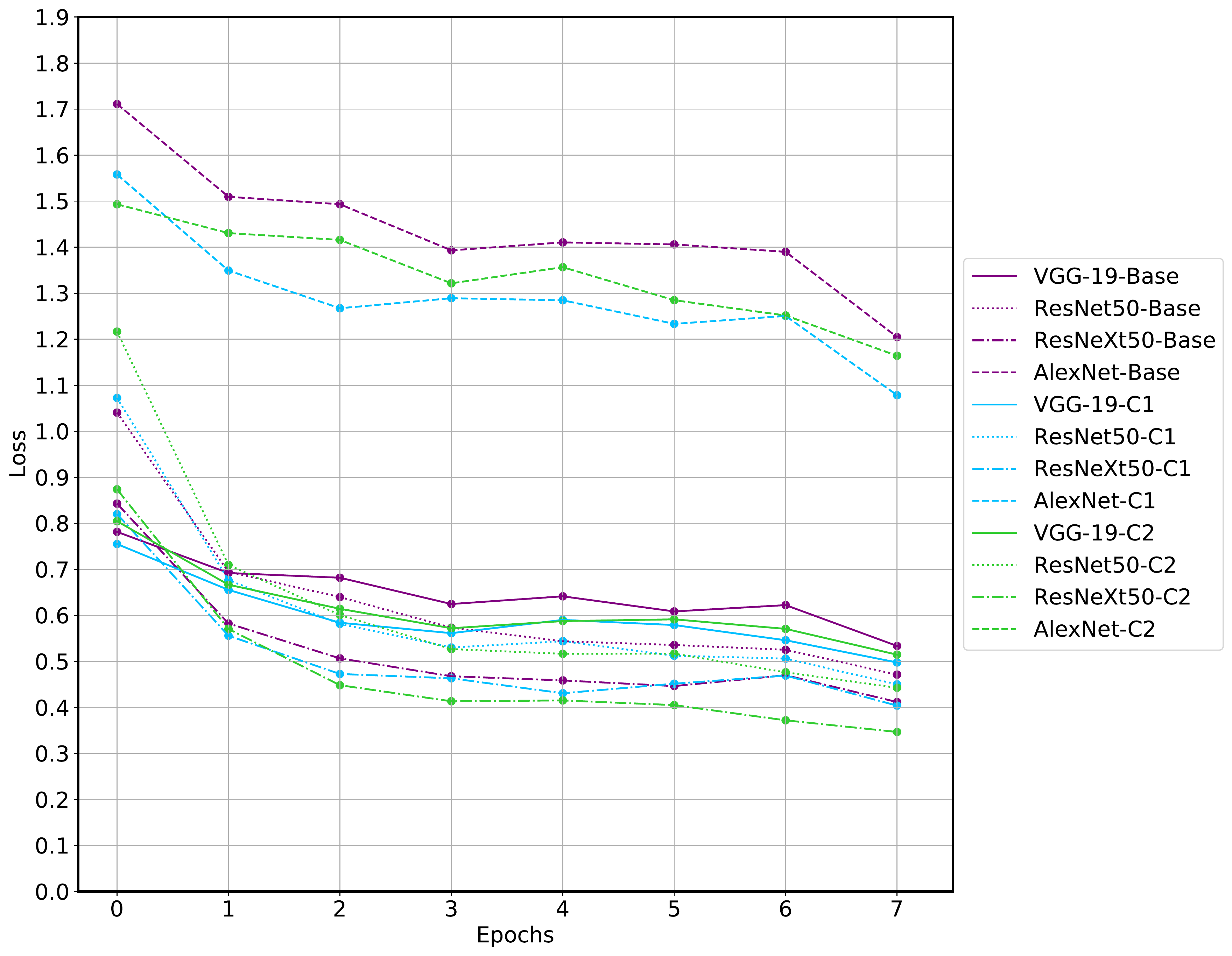}
    \caption{Loss for each epoch for model architectures trained on two cluster splits and original Stanford dogs dataset.}
    \label{figure5}
\end{minipage}
\end{figure}

\begin{figure}[!t]
\centering
\begin{minipage}{.5\textwidth}
    \renewcommand{\thefigure}{6}
    \centering
    \captionsetup{width = 0.9\columnwidth}
    \includegraphics[width = 0.9\columnwidth,scale=0.2]{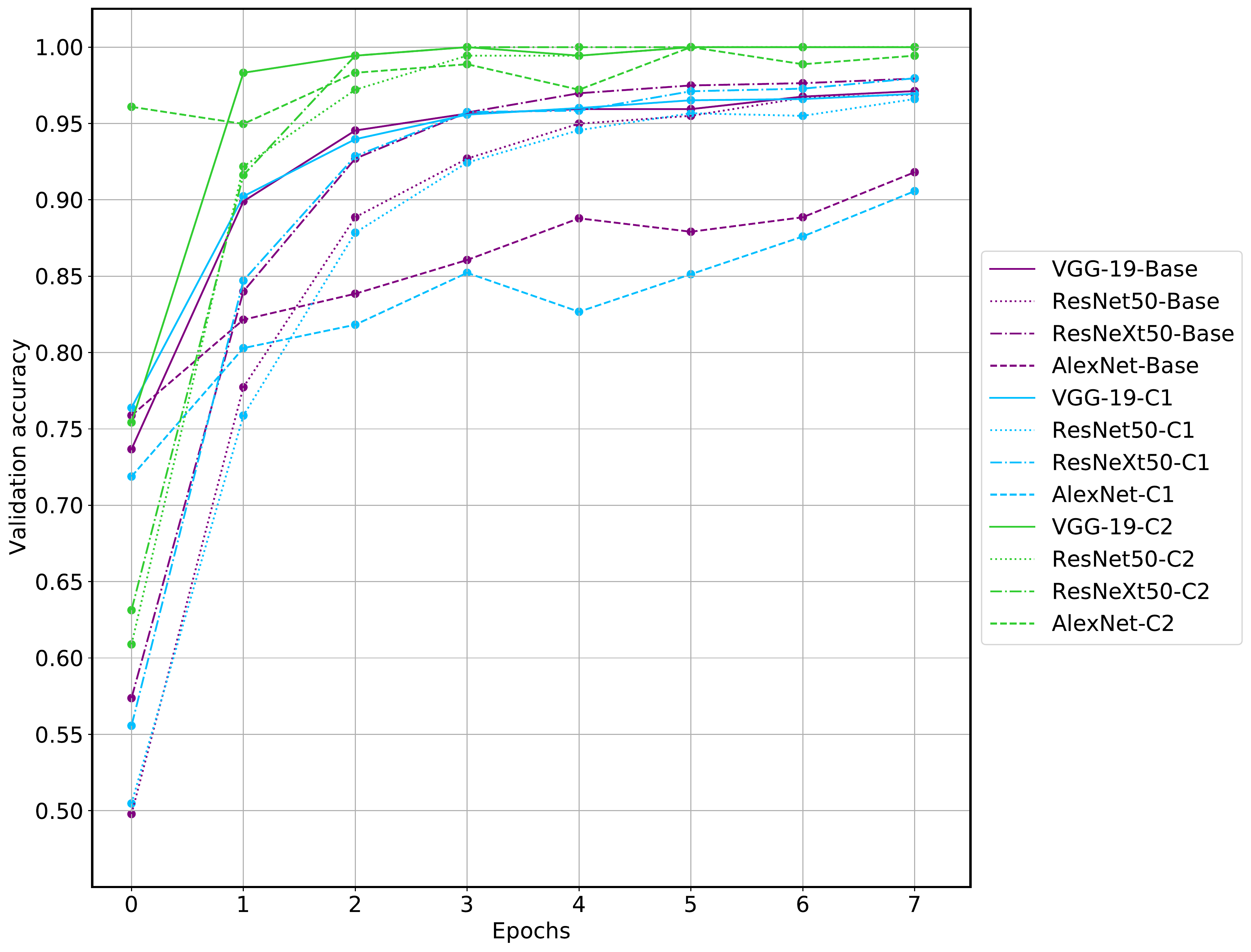}
    \caption{Validation accuracies for each epoch for model architectures trained on two cluster splits and original Oxford flowers dataset.}
    \label{figure8}
\end{minipage}%
\begin{minipage}{.5\textwidth}
    \renewcommand{\thefigure}{7}
    \centering
    \captionsetup{width = 0.8\columnwidth}
    \includegraphics[width = 0.9\columnwidth,scale=0.2]{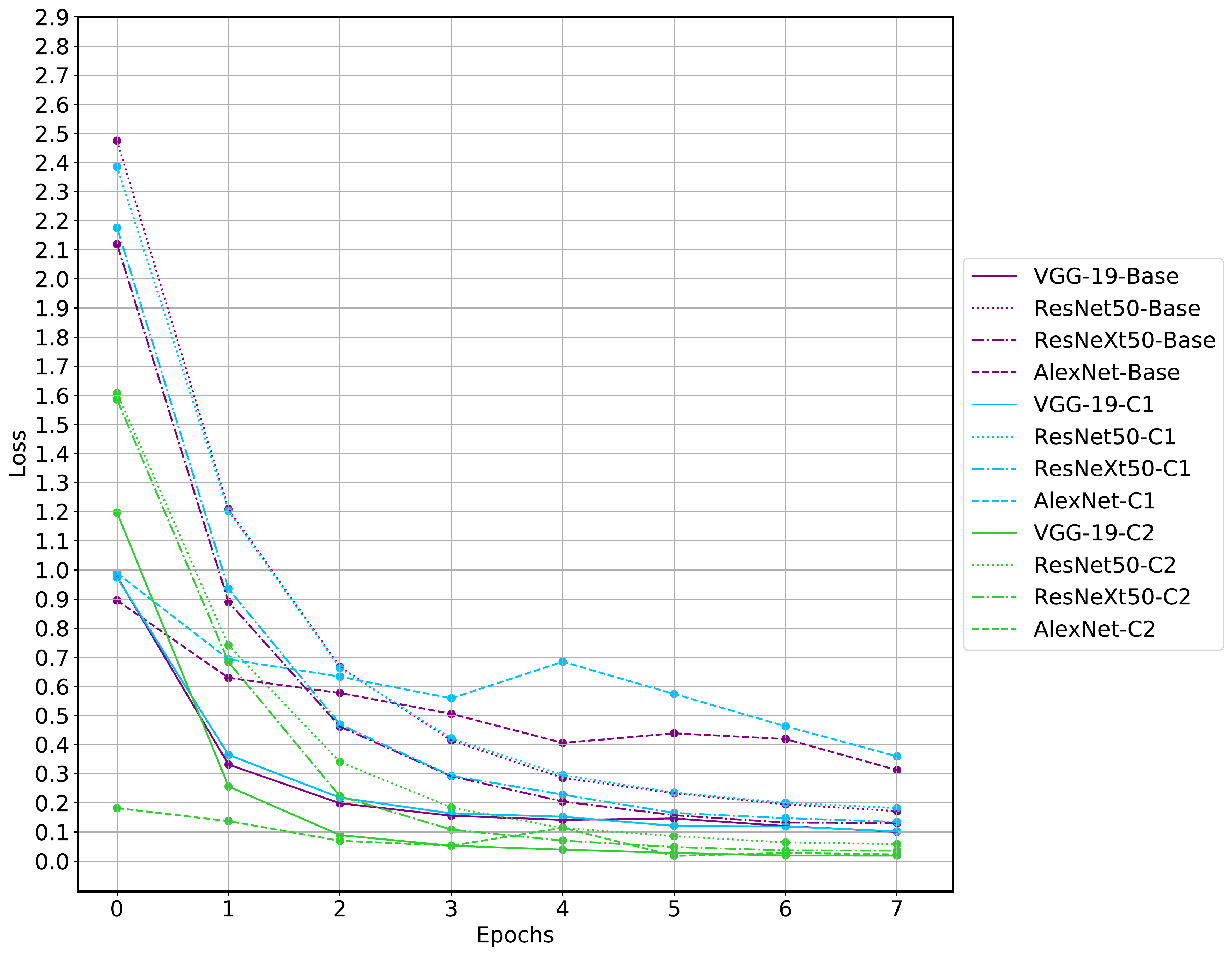}
    \caption{Loss for each epoch for model architectures trained on two cluster splits and original Oxford flowers dataset.}
    \label{figure9}
\end{minipage}
\end{figure}

In figure \ref{figure4} and \ref{figure5} (for Stanford dogs dataset), and \ref{figure8} and \ref{figure9} (for Oxford flowers dataset), we can see the trend of training in terms of validation accuracies and loss values for each epoch on all the model architectures trained on all three versions of the dataset. The trend for training is the same for all architectures and versions of the dataset. It shows that we can individually work upon a single cluster to quickly try and assess the model architectures and training pipelines. The concept of similarity-based clustering helps in improving the pipelines by taking the most difficult task aside and working on it. This also means that if the training pipeline works well on the single cluster, it will work well on the original dataset too. Hence, clustered training leads to two versions for inference, i.e., using the original dataset trained after bettering the model on the sub-dataset cluster and using the models trained on the clustered datasets directly, with the inference technique shown in Algorithm \ref{algo:algo3}. 

\begin{table}[ht]
\caption{Test accuracies of model architectures trained on different versions of Oxford Flowers dataset}
\begin{center}
\begin{tabularx}{\textwidth}{X X X X}
\hline
\textbf{Model architectures} & \textbf{Full dataset} & \textbf{Two cluster} & \textbf{Three cluster}\\ \hline
VGG-19 &         97.86 &        99.02 &          \textbf{99.04} \\
VGG-16 &         98.01 &        98.64 &          \textbf{98.68} \\
ResNet50 &         97.42 &        98.55 &          \textbf{98.89} \\
ResNext50 &         98.23 &        98.98 &          \textbf{99.05} \\
AlexNet &         92.55 &        95.71 &          \textbf{96.25} \\ \hline
\end{tabularx}
\label{table1}
\end{center}
\end{table}

\begin{table}[ht]
\caption{Test accuracies of model architectures trained on different versions of Stanford dogs dataset}
\begin{center}
\begin{tabularx}{\textwidth}{X X X X}
\hline
\textbf{Model architectures} & \textbf{Full dataset} & \textbf{Two cluster} & \textbf{Three cluster}\\ \hline
VGG-19 &         84.38 &        84.93 &          \textbf{85.97} \\
VGG-16 &         82.87 &        83.83 &          \textbf{84.74} \\
ResNet50 &         86.57 &        86.97 &          \textbf{87.54} \\
ResNext50 &         88.11 &        88.81 &          \textbf{89.63} \\
AlexNet &         66.46 &        67.97 &          \textbf{72.17} \\ \hline
\end{tabularx}
\label{table2}
\end{center}
\end{table}

In Table \ref{table1} we see test accuracies of the same model architectures on three different versions of the Oxford flowers dataset, i.e. original dataset, two-cluster dataset and three-cluster dataset. As observed, the latter two always out performed or were equal in comparison with training on the original dataset. Not just that but the same could be observed in Table \ref{table2}, where the dataset used is Stanford Dogs dataset. As mentioned earlier, all five model architectures were similar over all three types of dataset. Still, it either maintained the test accuracies or, though only by a small amount, increased it. 

When we train on sub-dataset clusters, the amount of computational resources required drops significantly. As the fundamental load of the amount of data is on runtime memory, the amount of memory required to cache the dataset is also reduced. It makes the learning possible even on lower computational resources. 

\renewcommand{\thefigure}{8}
\begin{figure}[!t]
\centering
\includegraphics[width = 0.9\columnwidth,scale=0.2]{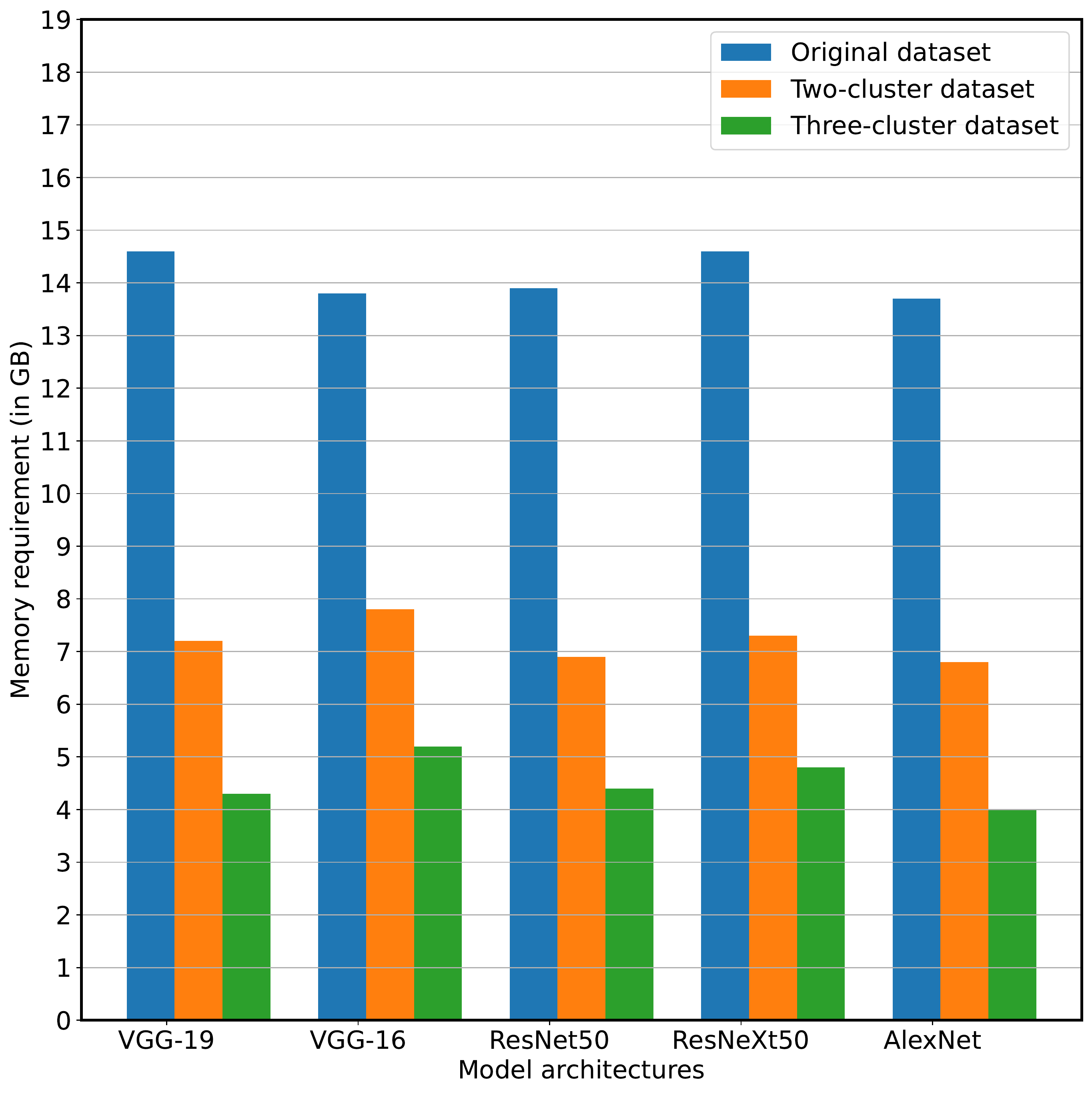}
\caption{Computational resources used for training.}
\label{figure6}
\end{figure}

In figure \ref{figure6}, we see the overall usage of model training resources required for training the model architectures on each version of the dataset. As we can see, all the model architectures, when trained on sub-dataset clusters, use a smaller amount of computational resources to cache the data, especially in terms of memory. It is mainly because of the amount of data reduced due to clustering the entire dataset into sub-datasets. Additionally, there is no additional toll on the model training timings. The total time for training remains the same. It makes it possible for parallelization of training as well. 

One of the most significant applications of similarity-based clustered training is the ability to have faster extensions. Let us take, for example, there is a model trained on 120 classes and now there is a need for appending 1 more class to the cluster. In terms of joint training, the model would need to train on 121 classes and hence would take a considerable amount of time. If the same model is trained on 4 sub-dataset clusters with 30 classes each, the new class can be associated with one of those clusters and retrained with only that cluster. So, the training would happen for only 31 classes, rather than 121 classes. Hence, there is an ability to optimize the speed and efficiency of small class extensions with sub-dataset clustering.

\section{Future works and conclusion}

The paper shows how the knowledge of similarity of the image classes can be leveraged in making the model training better and faster. It can also be used when the computational capacity is low for a single thread. The dataset can be broken down easily into clusters, which are trained individually, and then a rule-based system can be provided to make the appropriate model choice for predictions.
The paper shows that the methodology is adaptive, but still, more work can be done in improving the dynamic feature extractions as well as in reducing the error of choice. The clustering algorithms can be changed and chosen according to the way data is represented. Here we do leave the room for additional parameters on which the methodology can be improved, and comparative analysis with different clustering, number of clusters, and base model architectures can be performed. There is still the possibility of improving the way the clusters are formed or the way the similarity is extracted. The methodology, however, has to be considered. I wanted to drive the deep learning architectures and the modern datasets into the choice of declaring the similarity matrix, to have better knowledge as well as decision capacity of choosing the correct model for classification.

\section{Funding Source Declaration}
This research did not receive any specific grant from funding agencies in the public, commercial, or not-for-profit sectors.
\bibliographystyle{elsarticle-num}
\bibliography{elsarticle-template}

\begin{thebibliography}{10}
\expandafter\ifx\csname url\endcsname\relax
  \def\url#1{\texttt{#1}}\fi
\expandafter\ifx\csname urlprefix\endcsname\relax\def\urlprefix{URL }\fi
\expandafter\ifx\csname href\endcsname\relax
  \def\href#1#2{#2} \def\path#1{#1}\fi

\bibitem{huang2017densely}
G.~Huang, Z.~Liu, L.~Van Der~Maaten, K.~Q. Weinberger, Densely connected
  convolutional networks, in: Proceedings of the IEEE conference on computer
  vision and pattern recognition, 2017, pp. 4700--4708.

\bibitem{szegedy2017inception}
C.~Szegedy, S.~Ioffe, V.~Vanhoucke, A.~A. Alemi, Inception-v4, inception-resnet
  and the impact of residual connections on learning, in: Thirty-first AAAI
  conference on artificial intelligence, 2017.

\bibitem{he2016deep}
K.~He, X.~Zhang, S.~Ren, J.~Sun, Deep residual learning for image recognition,
  in: Proceedings of the IEEE conference on computer vision and pattern
  recognition, 2016, pp. 770--778.

\bibitem{howard2017mobilenets}
A.~G. Howard, M.~Zhu, B.~Chen, D.~Kalenichenko, W.~Wang, T.~Weyand,
  M.~Andreetto, H.~Adam, Mobilenets: Efficient convolutional neural networks
  for mobile vision applications, arXiv preprint arXiv:1704.04861.

\bibitem{sandler2018mobilenetv2}
M.~Sandler, A.~Howard, M.~Zhu, A.~Zhmoginov, L.-C. Chen, Mobilenetv2: Inverted
  residuals and linear bottlenecks, in: Proceedings of the IEEE conference on
  computer vision and pattern recognition, 2018, pp. 4510--4520.

\bibitem{tan2019efficientnet}
M.~Tan, Q.~V. Le, Efficientnet: Rethinking model scaling for convolutional
  neural networks, arXiv preprint arXiv:1905.11946.

\bibitem{xie2020self}
Q.~Xie, M.-T. Luong, E.~Hovy, Q.~V. Le, Self-training with noisy student
  improves imagenet classification, in: Proceedings of the IEEE/CVF Conference
  on Computer Vision and Pattern Recognition, 2020, pp. 10687--10698.

\bibitem{krizhevsky2012imagenet}
A.~Krizhevsky, I.~Sutskever, G.~E. Hinton, Imagenet classification with deep
  convolutional neural networks, Advances in neural information processing
  systems 25 (2012) 1097--1105.

\bibitem{bianco2018benchmark}
S.~Bianco, R.~Cadene, L.~Celona, P.~Napoletano, Benchmark analysis of
  representative deep neural network architectures, IEEE Access 6 (2018)
  64270--64277.

\bibitem{khosla2011novel}
A.~Khosla, N.~Jayadevaprakash, B.~Yao, F.-F. Li, Novel dataset for fine-grained
  image categorization: Stanford dogs, in: Proc. CVPR Workshop on Fine-Grained
  Visual Categorization (FGVC), Vol.~2, 2011.

\bibitem{wang2014learning}
J.~Wang, Y.~Song, T.~Leung, C.~Rosenberg, J.~Wang, J.~Philbin, B.~Chen, Y.~Wu,
  Learning fine-grained image similarity with deep ranking, in: Proceedings of
  the IEEE Conference on Computer Vision and Pattern Recognition, 2014, pp.
  1386--1393.

\bibitem{ciocca2001content}
G.~Ciocca, R.~Schettini, Content-based similarity retrieval of trademarks using
  relevance feedback, Pattern Recognition 34~(8) (2001) 1639--1655.

\bibitem{liu2007survey}
Y.~Liu, D.~Zhang, G.~Lu, W.-Y. Ma, A survey of content-based image retrieval
  with high-level semantics, Pattern recognition 40~(1) (2007) 262--282.

\bibitem{long2003fundamentals}
F.~Long, H.~Zhang, D.~D. Feng, Fundamentals of content-based image retrieval,
  in: Multimedia information retrieval and management, Springer, 2003, pp.
  1--26.

\bibitem{wan2014deep}
J.~Wan, D.~Wang, S.~C.~H. Hoi, P.~Wu, J.~Zhu, Y.~Zhang, J.~Li, Deep learning
  for content-based image retrieval: A comprehensive study, in: Proceedings of
  the 22nd ACM international conference on Multimedia, 2014, pp. 157--166.

\bibitem{srivastava2014dropout}
N.~Srivastava, G.~Hinton, A.~Krizhevsky, I.~Sutskever, R.~Salakhutdinov,
  Dropout: a simple way to prevent neural networks from overfitting, The
  journal of machine learning research 15~(1) (2014) 1929--1958.

\bibitem{lu2007survey}
D.~Lu, Q.~Weng, A survey of image classification methods and techniques for
  improving classification performance, International journal of Remote sensing
  28~(5) (2007) 823--870.

\bibitem{iandola2014densenet}
F.~Iandola, M.~Moskewicz, S.~Karayev, R.~Girshick, T.~Darrell, K.~Keutzer,
  Densenet: Implementing efficient convnet descriptor pyramids, arXiv preprint
  arXiv:1404.1869.

\bibitem{liu2020image}
J.-e. Liu, F.-P. An, Image classification algorithm based on deep
  learning-kernel function, Scientific programming 2020.

\bibitem{glorot2011deep}
X.~Glorot, A.~Bordes, Y.~Bengio, Deep sparse rectifier neural networks, in:
  Proceedings of the fourteenth international conference on artificial
  intelligence and statistics, JMLR Workshop and Conference Proceedings, 2011,
  pp. 315--323.

\bibitem{glorot2010understanding}
X.~Glorot, Y.~Bengio, Understanding the difficulty of training deep feedforward
  neural networks, in: Proceedings of the thirteenth international conference
  on artificial intelligence and statistics, JMLR Workshop and Conference
  Proceedings, 2010, pp. 249--256.

\bibitem{szegedy2016rethinking}
C.~Szegedy, V.~Vanhoucke, S.~Ioffe, J.~Shlens, Z.~Wojna, Rethinking the
  inception architecture for computer vision, in: Proceedings of the IEEE
  conference on computer vision and pattern recognition, 2016, pp. 2818--2826.

\bibitem{smeulders2000content}
A.~W. Smeulders, M.~Worring, S.~Santini, A.~Gupta, R.~Jain, Content-based image
  retrieval at the end of the early years, IEEE Transactions on pattern
  analysis and machine intelligence 22~(12) (2000) 1349--1380.

\bibitem{sampat2009complex}
M.~P. Sampat, Z.~Wang, S.~Gupta, A.~C. Bovik, M.~K. Markey, Complex wavelet
  structural similarity: A new image similarity index, IEEE transactions on
  image processing 18~(11) (2009) 2385--2401.

\bibitem{agarwal2015detection}
R.~Agarwal, A.~Shankhadhar, R.~K. Sagar, Detection of lung cancer using content
  based medical image retrieval, in: 2015 Fifth International Conference on
  Advanced Computing \& Communication Technologies, IEEE, 2015, pp. 48--52.

\bibitem{hoiem2004object}
D.~Hoiem, R.~Sukthankar, H.~Schneiderman, L.~Huston, Object-based image
  retrieval using the statistical structure of images, in: Proceedings of the
  2004 IEEE Computer Society Conference on Computer Vision and Pattern
  Recognition, 2004. CVPR 2004., Vol.~2, IEEE, 2004, pp. II--II.

\bibitem{sasaki2005formulation}
H.~Sasaki, Y.~Kiyoki, A formulation for patenting content-based retrieval
  processes in digital libraries, Information processing \& management 41~(1)
  (2005) 57--74.

\bibitem{nilsback2008automated}
M.-E. Nilsback, A.~Zisserman, Automated flower classification over a large
  number of classes, in: 2008 Sixth Indian Conference on Computer Vision,
  Graphics \& Image Processing, IEEE, 2008, pp. 722--729.

\bibitem{yosinski2014transferable}
J.~Yosinski, J.~Clune, Y.~Bengio, H.~Lipson, How transferable are features in
  deep neural networks?, arXiv preprint arXiv:1411.1792.

\bibitem{Pohchih2020}
Huang, Cbir system, \url{https://github.com/pochih/CBIR} (2020).

\bibitem{bressem2020comparing}
K.~K. Bressem, L.~C. Adams, C.~Erxleben, B.~Hamm, S.~M. Niehues, J.~L.
  Vahldiek, Comparing different deep learning architectures for classification
  of chest radiographs, Scientific reports 10~(1) (2020) 1--16.

\bibitem{hasnat2013comparative}
A.~Hasnat, S.~Halder, D.~Bhattacharjee, M.~Nasipuri, D.~Basu, Comparative study
  of distance metrics for finding skin color similarity of two color facial
  images, ACER: New Taipei City, Taiwan (2013) 99--108.

\bibitem{vijaya2017review}
A.~S. Vijaya, R.~Bateja, A review on hierarchical clustering algorithms, J.
  Eng. Appl. Sci 12~(24) (2017) 7501--7507.

\end{thebibliography}

\appendix \pagebreak
\section{Two and three cluster splits - Stanford Dog Breeds}
\label{appendix:a}

\begin{longtable}{llr}
\toprule
                         Breed & Cluster in Two-splits & Cluster in three-splits \\
\midrule
                Scotch\_terrier &                     1 &                       1 \\
        Coated\_wheaten\_terrier &                     1 &                       1 \\
                            Lh &                     1 &                       1 \\
                      Papillon &                     1 &                       1 \\
                   Maltese\_dog &                     1 &                       1 \\
                      Chihuahu &                     2 &                       2 \\
                        Redbon &                     2 &                       2 \\
                           Tzu &                     1 &                       1 \\
               Norfolk\_terrier &                     1 &                       1 \\
                      Bluetick &                     2 &                       2 \\
            Australian\_terrier &                     1 &                       1 \\
              Italian\_greyhoun &                     2 &                       2 \\
      Chesapeake\_bay\_retriever &                     2 &                       2 \\
                   Afghan\_houn &                     1 &                       1 \\
            Labrador\_retriever &                     2 &                       2 \\
                 Irish\_terrier &                     1 &                       1 \\
               Giant\_schnauzer &                     1 &                       1 \\
                         Whipp &                     2 &                       2 \\
              Coated\_retriever &                     1 &                       1 \\
           Rhodesian\_ridgeback &                     2 &                       2 \\
             Scottish\_deerhoun &                     1 &                       1 \\
              Coated\_retriever &                     1 &                       1 \\
               English\_foxhoun &                     2 &                       2 \\
               Tibetan\_terrier &                     1 &                       1 \\
               Norwich\_terrier &                     1 &                       1 \\
                     Otterhoun &                     1 &                       1 \\
            Standard\_schnauzer &                     1 &                       1 \\
                  Tan\_coonhoun &                     2 &                       2 \\
                               &                     2 &                       2 \\
                        Borzoi &                     1 &                       1 \\
                         Cairn &                     1 &                       1 \\
            Bedlington\_terrier &                     1 &                       1 \\
              Japanese\_spaniel &                     1 &                       1 \\
                     Bloodhoun &                     2 &                       2 \\
           Miniature\_schnauzer &                     1 &                       1 \\
                 Dandie\_dinmon &                     1 &                       1 \\
            Haired\_fox\_terrier &                     1 &                       1 \\
                         Beagl &                     2 &                       2 \\
             Yorkshire\_terrier &                     1 &                       1 \\
                   Walker\_houn &                     2 &                       2 \\
                Irish\_wolfhoun &                     1 &                       1 \\
              Golden\_retriever &                     1 &                       1 \\
                         Pekin &                     1 &                       1 \\
              Sealyham\_terrier &                     1 &                       1 \\
                    Weimaraner &                     2 &                       2 \\
     Staffordshire\_bullterrier &                     2 &                       2 \\
                Border\_terrier &                     1 &                       1 \\
            Kerry\_blue\_terrier &                     1 &                       1 \\
             Norwegian\_elkhoun &                     1 &                       3 \\
                       Airedal &                     1 &                       1 \\
                        Saluki &                     1 &                       1 \\
              Lakeland\_terrier &                     1 &                       1 \\
   West\_highland\_white\_terrier &                     1 &                       1 \\
                Haired\_pointer &                     2 &                       2 \\
              Blenheim\_spaniel &                     1 &                       1 \\
                   Ibizan\_houn &                     2 &                       2 \\
                   Toy\_terrier &                     2 &                       2 \\
American\_staffordshire\_terrier &                     2 &                       2 \\
                   Boston\_bull &                     2 &                       2 \\
                 Silky\_terrier &                     1 &                       1 \\
                     Great\_dan &                     2 &                       2 \\
                  Bull\_mastiff &                     2 &                       2 \\
           African\_hunting\_dog &                     2 &                       2 \\
                         Boxer &                     2 &                       2 \\
                   Great\_pyren &                     1 &                       1 \\
                German\_shepher &                     2 &                       2 \\
          Bernese\_mountain\_dog &                     1 &                       1 \\
                      Doberman &                     2 &                       2 \\
                       Keeshon &                     1 &                       3 \\
            Miniature\_pinscher &                     2 &                       2 \\
                      Cardigan &                     2 &                       2 \\
          Old\_english\_sheepdog &                     1 &                       1 \\
             Brabancon\_griffon &                     2 &                       2 \\
             Shetland\_sheepdog &                     1 &                       1 \\
                 Affenpinscher &                     1 &                       1 \\
                    Pomeranian &                     1 &                       3 \\
                   Groenendael &                     1 &                       3 \\
               Tibetan\_mastiff &                     1 &                       1 \\
                        Malamu &                     1 &                       3 \\
                       Clumber &                     1 &                       1 \\
                   Appenzeller &                     2 &                       2 \\
                English\_setter &                     1 &                       1 \\
                Cocker\_spaniel &                     1 &                       1 \\
        Welsh\_springer\_spaniel &                     1 &                       1 \\
                     Toy\_poodl &                     1 &                       1 \\
                          Dhol &                     2 &                       2 \\
                Siberian\_husky &                     1 &                       3 \\
                  Irish\_setter &                     1 &                       1 \\
                        Kuvasz &                     1 &                       1 \\
    Greater\_swiss\_mountain\_dog &                     2 &                       2 \\
                       Pembrok &                     2 &                       2 \\
              English\_springer &                     1 &                       1 \\
                Sussex\_spaniel &                     1 &                       1 \\
                  Border\_colli &                     1 &                       1 \\
            Bouvier\_des\_flandr &                     1 &                       1 \\
                 Mexican\_hairl &                     2 &                       2 \\
                         Dingo &                     2 &                       2 \\
                Standard\_poodl &                     1 &                       1 \\
                     Schipperk &                     1 &                       3 \\
                  Saint\_bernar &                     1 &                       1 \\
                   Entlebucher &                     2 &                       2 \\
                       Malinoi &                     2 &                       2 \\
                         Vizsl &                     2 &                       2 \\
               Miniature\_poodl &                     1 &                       1 \\
                         Briar &                     1 &                       1 \\
                    Rottweiler &                     2 &                       2 \\
                         Samoy &                     1 &                       3 \\
                       Basenji &                     2 &                       2 \\
                   Newfoundlan &                     1 &                       1 \\
                           Pug &                     2 &                       2 \\
              Brittany\_spaniel &                     1 &                       1 \\
                      Komondor &                     1 &                       1 \\
                    Eskimo\_dog &                     1 &                       3 \\
                         Colli &                     1 &                       1 \\
                      Leonberg &                     1 &                       1 \\
                 Gordon\_setter &                     1 &                       1 \\
                         Kelpi &                     2 &                       2 \\
                French\_bulldog &                     2 &                       2 \\
           Irish\_water\_spaniel &                     1 &                       1 \\
                          Chow &                     1 &                       3 \\
\bottomrule
\end{longtable}

\section{Two and three cluster splits - Oxford flowers}
\label{appendix:b}

\begin{longtable}{llr}

\toprule
                  Species & Cluster in Two-splits & Cluster in three-splits \\
\midrule
            Pink primrose &                     2 &                       2 \\
            Globe thistle &                     1 &                       1 \\
           Blanket flower &                     2 &                       2 \\
          Trumpet creeper &                     2 &                       2 \\
          Blackberry lily &                     1 &                       1 \\
               Snapdragon &                     2 &                       2 \\
              Colt's foot &                     1 &                       1 \\
              King protea &                     2 &                       2 \\
            Spear thistle &                     1 &                       1 \\
              Yellow iris &                     2 &                       2 \\
             Globe-flower &                     2 &                       2 \\
        Purple coneflower &                     2 &                       2 \\
            Peruvian lily &                     2 &                       2 \\
           Balloon flower &                     2 &                       2 \\
Hard-leaved pocket orchid &                     2 &                       3 \\
    Giant white arum lily &                     2 &                       3 \\
                Fire lily &                     2 &                       2 \\
        Pincushion flower &                     2 &                       2 \\
               Fritillary &                     1 &                       1 \\
               Red ginger &                     2 &                       2 \\
           Grape hyacinth &                     1 &                       1 \\
               Corn poppy &                     2 &                       2 \\
 Prince of wales feathers &                     2 &                       3 \\
         Stemless gentian &                     2 &                       2 \\
                Artichoke &                     2 &                       3 \\
         Canterbury bells &                     2 &                       2 \\
            Sweet william &                     2 &                       3 \\
                Carnation &                     2 &                       2 \\
             Garden phlox &                     2 &                       3 \\
         Love in the mist &                     1 &                       1 \\
            Mexican aster &                     2 &                       2 \\
         Alpine sea holly &                     1 &                       1 \\
     Ruby-lipped cattleya &                     2 &                       2 \\
              Cape flower &                     2 &                       2 \\
         Great masterwort &                     2 &                       2 \\
               Siam tulip &                     2 &                       2 \\
                Sweet pea &                     2 &                       2 \\
              Lenten rose &                     2 &                       2 \\
           Barbeton daisy &                     2 &                       2 \\
                 Daffodil &                     2 &                       2 \\
               Sword lily &                     2 &                       2 \\
               Poinsettia &                     2 &                       3 \\
         Bolero deep blue &                     2 &                       2 \\
               Wallflower &                     1 &                       1 \\
                 Marigold &                     2 &                       3 \\
                Buttercup &                     2 &                       2 \\
              Oxeye daisy &                     1 &                       1 \\
         English marigold &                     2 &                       2 \\
         Common dandelion &                     1 &                       1 \\
                  Petunia &                     2 &                       2 \\
               Wild pansy &                     2 &                       2 \\
                  Primula &                     2 &                       3 \\
                Sunflower &                     2 &                       2 \\
              Pelargonium &                     2 &                       2 \\
       Bishop of llandaff &                     2 &                       2 \\
                    Gaura &                     1 &                       1 \\
                 Geranium &                     2 &                       3 \\
            Orange dahlia &                     2 &                       2 \\
               Tiger lily &                     2 &                       2 \\
       Pink-yellow dahlia &                     2 &                       3 \\
         Cautleya spicata &                     2 &                       2 \\
         Japanese anemone &                     2 &                       2 \\
         Black-eyed susan &                     1 &                       1 \\
               Silverbush &                     2 &                       3 \\
        Californian poppy &                     2 &                       2 \\
             Osteospermum &                     1 &                       1 \\
            Spring crocus &                     2 &                       2 \\
             Bearded iris &                     2 &                       3 \\
               Windflower &                     2 &                       2 \\
              Moon orchid &                     2 &                       3 \\
               Tree poppy &                     2 &                       3 \\
                  Gazania &                     2 &                       2 \\
                   Azalea &                     2 &                       2 \\
               Water lily &                     2 &                       2 \\
                     Rose &                     2 &                       2 \\
              Thorn apple &                     2 &                       3 \\
            Morning glory &                     2 &                       2 \\
           Passion flower &                     2 &                       2 \\
                    Lotus &                     2 &                       2 \\
                Toad lily &                     1 &                       1 \\
         Bird of paradise &                     2 &                       2 \\
                Anthurium &                     2 &                       2 \\
               Frangipani &                     2 &                       2 \\
                 Clematis &                     2 &                       2 \\
                 Hibiscus &                     2 &                       2 \\
                Columbine &                     2 &                       2 \\
              Desert-rose &                     2 &                       2 \\
              Tree mallow &                     2 &                       2 \\
                 Magnolia &                     2 &                       2 \\
                 Cyclamen &                     2 &                       2 \\
               Watercress &                     2 &                       2 \\
                Monkshood &                     2 &                       2 \\
               Canna lily &                     2 &                       2 \\
              Hippeastrum &                     2 &                       2 \\
                 Bee balm &                     1 &                       1 \\
                Ball moss &                     2 &                       2 \\
                 Foxglove &                     2 &                       2 \\
            Bougainvillea &                     2 &                       2 \\
                 Camellia &                     2 &                       2 \\
                   Mallow &                     2 &                       2 \\
          Mexican petunia &                     2 &                       2 \\
                 Bromelia &                     2 &                       3 \\
\bottomrule
\end{longtable}

\end{document}